%% file: main.tex
\documentclass[final]{cvpr}
\usepackage{times}
\usepackage{epsfig}
\usepackage{graphicx}
\usepackage{amsmath}
\usepackage{amssymb}
\usepackage{bbold}
\usepackage{booktabs}
\usepackage{mathtools}
\usepackage{multirow}
\usepackage{pifont}
\usepackage{adjustbox}
\usepackage[caption=false]{subfig}
\usepackage{times}
\usepackage{xcolor}
\usepackage{algorithm}
\usepackage{listings}
\newcommand{\xmark}{\ding{55}}

\usepackage{etoolbox} 
\makeatletter
\AfterEndEnvironment{algorithm}{\let\@algcomment\relax}
\AtEndEnvironment{algorithm}{\kern2pt\hrule\relax\vskip3pt\@algcomment}
\let\@algcomment\relax
\newcommand\algcomment[1]{\def\@algcomment{\footnotesize#1}}
\renewcommand\fs@ruled{\def\@fs@cfont{\bfseries}\let\@fs@capt\floatc@ruled
  \def\@fs@pre{\hrule height.8pt depth0pt \kern2pt}%
  \def\@fs@post{}%
  \def\@fs@mid{\kern2pt\hrule\kern2pt}%
  \let\@fs@iftopcapt\iftrue}
\makeatother

\definecolor{Highlight}{HTML}{39b54a}  

\usepackage[pagebackref=true,breaklinks=true,colorlinks,bookmarks=false]{hyperref}

\newcommand{\app}{\raise.17ex\hbox{$\scriptstyle\sim$}}
\newcommand*\samethanks[1][\value{footnote}]{\footnotemark[#1]}

\begin{document}

\title{Unsupervised Semantic Segmentation by Contrasting Object Mask Proposals}

\author{Wouter Van Gansbeke$^1$\thanks{Authors contributed equally} \quad Simon Vandenhende$^1$\samethanks[1] \quad Stamatios Georgoulis$^2$ \quad Luc Van Gool$^{1,2}$ \\
$^1$KU Leuven/ESAT-PSI \quad $^2$ETH Zurich/CVL, TRACE}

\maketitle

\begin{abstract}
Being able to learn dense semantic representations of images without supervision is an important problem in computer vision. However, despite its significance, this problem remains rather unexplored, with a few exceptions that considered unsupervised semantic segmentation on small-scale datasets with a narrow visual domain. In this paper, we make a first attempt to tackle the problem on datasets that have been traditionally utilized for the supervised case. To achieve this, we introduce a two-step framework that adopts a predetermined mid-level prior in a contrastive optimization objective to learn pixel embeddings. This marks a large deviation from existing works that relied on proxy tasks or end-to-end clustering. Additionally, we argue about the importance of having a prior that contains information about objects, or their parts, and discuss several possibilities to obtain such a prior in an unsupervised manner.
   
Experimental evaluation shows that our method comes with key advantages over existing works. First, the learned pixel embeddings can be directly clustered in semantic groups using K-Means on PASCAL. Under the fully unsupervised setting, there is no precedent in solving the semantic segmentation task on such a challenging benchmark. Second, our representations can improve over strong baselines when transferred to new datasets, e.g. COCO and DAVIS. The code is available\footnote{{github.com/wvangansbeke/Unsupervised-Semantic-Segmentation.git}}.
\end{abstract}

\input{sections/introduction}
\input{sections/related_work}
\input{sections/method}

\input{sections/experiments}

\input{sections/conclusion}
\paragraph{Acknowledgment.}
The authors thankfully acknowledge support by Toyota via the TRACE project and MACCHINA (KU Leuven, C14/18/065).

\setcounter{section}{0}
\renewcommand\thesection{\Alph{section}}
\setcounter{figure}{0}
\setcounter{table}{0}
\renewcommand{\thefigure}{S\arabic{figure}}
\renewcommand{\thetable}{S\arabic{table}}
\section{Supplementary Materials}
\input{supp}

{\small
\bibliographystyle{ieee_fullname}
\bibliography{egbib}
}

\end{document}

%% file: sections/introduction.tex
\section{Introduction}
\label{sec: introduction}

\begin{figure}
    \centering
    \includegraphics[width=1.0\linewidth]{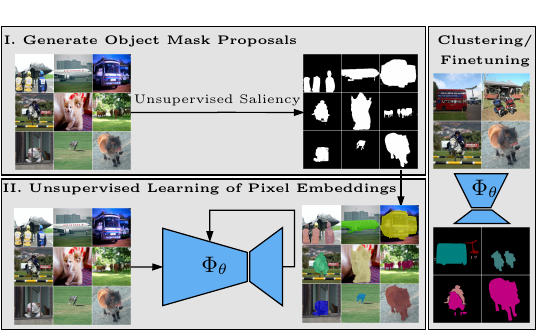}
    \caption{We learn pixel embeddings for semantic segmentation in an unsupervised way. First, we predict object mask proposals using unsupervised saliency. Second, we use the obtained masks as a prior in a self-supervised optimization objective. Finally, the pixel embeddings can be clustered or fine-tuned to a semantic segmentation of the image.}
    \label{fig: teaser}
\end{figure}

The problem of assigning dense semantic labels to images, formally known as \textit{semantic segmentation}, is of great importance in computer vision as it finds many applications, including autonomous driving, augmented reality, human-computer interaction, etc. To achieve state-of-the-art performance in this task, fully convolutional networks~\cite{long2015fully} are typically trained on datasets~\cite{cordts2016cityscapes,everingham2010pascal,lin2014microsoft} that contain a large number of fully-annotated images. However, obtaining accurate, pixel-wise semantic labels for every image in a dataset is a labor-intensive process that costs significant amounts of time and money~\cite{bearman2016whats}. To remedy this situation, weakly-supervised methods leveraged weaker forms of supervision, such as scribbles~\cite{lin2016scribblesup,tang2018normalized,tang2018regularized,vernaza2017learning,xu2015learning}, bounding boxes~\cite{dai2015boxsup,khoreva2017simple,papandreou2015weakly,xu2015learning}, clicks~\cite{bearman2016s}, and image-level tags~\cite{papandreou2015weakly,tang2018regularized,xu2015learning}, while semi-supervised methods~\cite{dai2015boxsup,hong2015decoupled,hung2018adversarial,papandreou2015weakly,pathak2015constrained} used only a fraction of the dataset as labeled examples, both of which require substantially less human annotation effort. Despite the continued progress, the vast majority of semantic segmentation works still rely on some form of annotations to train the neural network models.

In this paper, we look at the problem from a different perspective, namely self-supervised representation learning. More concretely, we aim to learn pixel-level representations or embeddings for semantic segmentation without using ground-truth. If we obtain a good pixel embedding~\cite{de2017semantic} that is discriminative w.r.t. the semantic classes, we can directly cluster the pixels into semantic groups using K-Means. This tackles the semantic segmentation problem under the fully unsupervised setup. Alternatively, if a limited number of annotated examples are available, the representations can be further fine-tuned under a semi-supervised or transfer learning setup. In this paper, we primarily focus on the fully unsupervised setup, but include additional fine-tuning experiments for the sake of completeness.  


Unsupervised or self-supervised techniques~\cite{jing2020self} were recently being employed to learn rich and effective visual representations without external supervision. The obtained representations can subsequently be used for a variety of purposes, including task transfer learning~\cite{he2019momentum}, image clustering~\cite{asano2020labelling,asano20self,van2020scan}, semi-supervised classification~\cite{chen2020big}, etc. Popular representation learning techniques used an instance discrimination task~\cite{wu2018unsupervised}, that is treating every image as a separate class, to generate representations in an unsupervised way. Images and their augmentations are considered as positive examples of the class, while all other images are treated as negatives. In practical terms, the instance discrimination task is formulated as a non-parametric classification problem, and a contrastive loss~\cite{gutmann2010noise,oord2018representation} is used to model the distribution of negative instance classes.

Purushwalkam and Gupta~\cite{purushwalkam2020demystifying} showed that contrastive self-supervised methods learn to encode semantic information, since two views of the same image will always show a part of the same object, and no objects from other categories. However, under this setup, there is no guarantee that the representations also learn to differentiate between pixels belonging to different semantic classes. For example, when foreground-background pairs frequently co-occur, e.g. cattle grazing on farmland, pixels belonging to the two classes can share their representation. This renders existing works based on instance discrimination less appropriate w.r.t. our goal of learning semantic pixel embeddings. To address these limitations, we propose to learn pixel-level, rather than image-level representations, in a self-supervised way.

The proposed method consists of two steps. First, we leverage an unsupervised saliency estimator to mine object mask proposals from the dataset. This mid-level visual prior transfers well across different datasets. In the second step, we use a contrastive framework to learn pixel embeddings. The object mask proposals are employed as a prior - we pull embeddings from pixels belonging to the same object together, and contrast them against pixels from other objects. The generated representations are evaluated on the semantic segmentation task following standard protocols. The framework is illustrated in Figure~\ref{fig: teaser}. 

Our contributions are: (1) We propose a two-step framework for unsupervised semantic segmentation, which marks a large deviation from recent works that relied on proxy tasks or end-to-end clustering. Additionally, we argue about the importance of having a mid-level visual prior which incorporates object-level information. This contrasts with earlier works that grouped pixels together based upon low-level vision tasks like boundary detection. (2) The proposed method is the first able to tackle the semantic segmentation task on a challenging dataset like PASCAL under the fully unsupervised setting. (3) Finally, we report promising results when transferring our representations to other datasets. This shows that adopting a mid-level visual prior can be useful for self-supervised representation learning. 

%% file: sections/related_work.tex
\section{Related Work}
\label{sec: related_work}
As our method is mostly related to unsupervised semantic segmentation and representation learning, in what follows we discuss representative works from each topic. 

\paragraph{Unsupervised semantic segmentation.} There have only been a few attempts in the literature to tackle semantic image segmentation under the fully unsupervised setting. Some works~\cite{xu2019invariant,ouali2020autoregressive} followed an end-to-end approach - maximizing the discrete mutual information between augmented views to learn a clustering function. However, these methods were only applied to small-scale datasets, covering a narrow visual domain, e.g. separating sky from vegetation, using satellite imagery, etc. In contrast, our method applies to more challenging scenarios, and decouples feature learning from clustering.

A few works~\cite{hwang2019segsort,zhang2020self} used segments obtained from boundaries to learn pixel embeddings in a self-supervised way. However, it is unclear whether the representations could be post-processed with an off-line clustering criterion to obtain discrete labels. In particular, the evaluation only considered semantic segment retrieval which requires an annotated train set. Furthermore, Hwang~\etal~\cite{hwang2019segsort} still relied on additional supervision sources like ImageNet pre-training and boundary annotations~\cite{arbelaez2010contour,xie2015holistically}.

\paragraph{Representation learning.} These methods aim at learning visual representations by solving pre-designed \emph{pretext tasks}, which do not require manual annotations. Examples of such pretext tasks include colorizing images~\cite{iizuka2016let,larsson2017colorization,zhang2016colorful}, predicting context~\cite{doersch2015unsupervised,nathan2018improvements}, solving jigsaw puzzles~\cite{noroozi2016unsupervised,noroozi2018boosting}, generating images~\cite{ren2018cross}, clustering~\cite{asano2020labelling,caron2018deep,yan2020clusterfit}, predicting noise~\cite{bojanowski2017unsupervised}, spotting artifacts~\cite{jenni2018self}, using adversarial training~\cite{donahue2017adversarial,donahue2019large}, predicting optical flow~\cite{mahendran2018cross,zhan2019self}, counting~\cite{noroozi2017representation}, inpainting~\cite{pathak2016context}, predicting transformation parameters~\cite{gidaris2018unsupervised,zhang2019aet}, using predictive coding~\cite{oord2018representation}, performing instance discrimination~\cite{caron2020unsupervised,chaitanya2020contrastive,chen2020simple,grill2020bootstrap,he2019momentum,li2020prototypical,misra2020self,tian2019contrastive,tian2020infomin,wu2018unsupervised,ye2019unsupervised}, and so on. The learned representations can subsequently be transferred to learn a separate down-stream task, e.g. object detection.

In a similar vein, some works tried to learn pixel-level representations for semantic segmentation by solving proxy tasks, e.g. colorization~\cite{iizuka2016let,larsson2017colorization,zhan2018mix,zhang2016colorful}, optical flow~\cite{mahendran2018cross,zhan2019self}, using co-occurences~\cite{isola2015learning}, etc. Differently, in this paper, we avoid the use of a proxy task.

%% file: sections/method.tex
\section{Method}
\label{sec: method}

\begin{figure}
    \centering
    \includegraphics[width=1.0\linewidth]{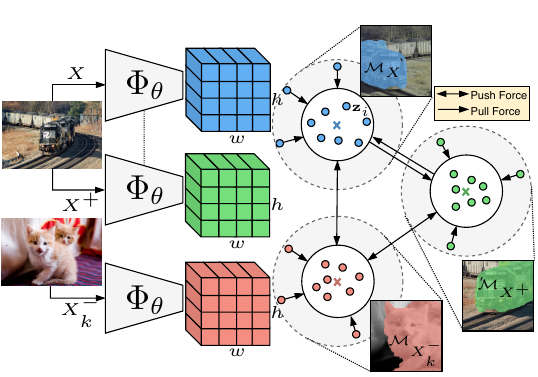}
    \caption{\textbf{MaskContrast} learns pixel embeddings for unsupervised semantic segmentation in the following way. We use a saliency estimator to generate positive pairs of object-centric crops $(X, X^+)$ and negative pairs $X^-_k$. The model $\Phi_\theta$ is trained to maximize the agreement between embeddings of pixels belonging to the objects in $X, X^+$, while minimizing the agreement with pixels from objects in $X^-_k$.}
    \label{fig: pipeline}
\end{figure}

In this paper, we aim to learn a pixel embedding function for semantic segmentation from an unlabeled dataset of images. Since the goal of semantic segmentation is to assign a class label to every pixel of an image, a good pixel embedding should be discriminative w.r.t. the semantic classes. If the latter holds true, the embedding function can be directly used to cluster the pixels into semantic groups, or be further fine-tuned under a semi-supervised setup. 

To tackle the aforementioned problem, we follow a divide-and-conquer strategy. We argue that it is more difficult to directly cluster the pixels into semantic groups following an end-to-end pipeline, while it is easier to first look for image regions where pixels are likely to belong together. Although this information does not directly result in a semantic segmentation of the scene, it gives us a useful starting point to learn the pixel embeddings. In particular, we can leverage the obtained regions as a prior by grouping their pixels together. Since the prior is determined before the feature learning step, we reduce the dependence on the network initialization. This is an intentional divergence from existing end-to-end learning pipelines~\cite{xu2019invariant,ouali2020autoregressive}, which are prone to latch onto low-level image cues - like color, contrast, etc. - as shown in~\cite{van2020scan}.

The proposed method named \emph{MaskContrast} consists of two steps. In a first step, we determine a prior by identifying objects in the images for which pixels can be grouped together. Mid-level visual groups, like objects, transfer well across datasets, since they do not depend on any pre-defined ground-truth classes. In the second step, we employ the obtained prior in a contrastive loss~\cite{gutmann2010noise,oord2018representation} to generate pixel embeddings. More specifically, we pull pixels belonging to the same object together, and contrast them against pixels from other objects, as shown in Figure~\ref{fig: pipeline}. This forces the model to map pixels from visually similar objects closer together, while pushing pixels from dissimilar objects further apart. In this way, the model discovers a pixel embedding space that can serve as a dense semantic representation of the scene. 

The method section is further organized as follows. Section~\ref{subsec: prior} motivates the use of object mask proposals as a prior for semantic segmentation. Section~\ref{subsec: saliency} analyzes the use of an unsupervised saliency estimator to mine the object masks from unlabeled datasets. Section~\ref{subsec: self_supervised} integrates the prior in a contrastive loss to learn pixel embeddings. 

\subsection{A Mid-Level Visual Prior for Grouping Pixels}
\label{subsec: prior}
As a starting point for unsupervised semantic segmentation, we try to define an appropriate prior. Several works have emerged in the literature that tried to group pixels by solving a proxy task. Examples include colorizing images~\cite{iizuka2016let,larsson2017colorization,zhang2016colorful}, predicting optical flow~\cite{mahendran2018cross,zhan2019self}, using co-occurences~\cite{isola2015learning}, etc. Unfortunately, there is no guarantee that the generated representations will align with the semantic classes, as the latter are co-variant to the proxy task's output. For example, a colorization network will be sensitive to color changes, even though these do not necessarily alter the semantics of the scene. This behavior is unwanted for the objective of semantic segmentation.

To overcome these limitations, we follow an alternative route that avoids the use of a proxy task. In particular, we mine object mask proposals which cover patches that are likely to contain an object. A prior can then be defined from the masks based upon \emph{shared pixel ownership}, i.e. if a pair of pixels belongs to the same mask, we assume that they should be grouped together, and maximize the agreement between their pixel embeddings. We hypothesize that this is a more reliable pixel grouping strategy compared to the use of proxy tasks. In particular, our approach builds a high-level image segmentation by first identifying mid-level visual groups, instead of directly producing a complete segmentation by solving a proxy task. A motivation for this bottom-up approach is also provided in~\cite{shi2000normalized}. 

At the same time, the proposed prior can be seen as an object-centric approach to unsupervised semantic segmentation, which brings several advantages to the table. First, using mid-level visual cues, like object information, regularizes the feature representations. In particular, the model can not simply rely on low-level information like color to group the pixels together, but needs to learn more semantically meaningful image characteristics. This differs from competing works~\cite{hwang2019segsort,zhang2020self} that used superpixels or image boundaries as a prior. Second, object cues can be highly informative of the semantic segmentation task. Evidence for the latter has been provided in the literature for weakly-supervised methods that utilized annotations containing object information. As an example, several works~\cite{dai2015boxsup,khoreva2017simple,papandreou2015weakly,xu2015learning} reported strong results on the segmentation task by employing object bounding boxes.

Next, we show how an unsupervised saliency estimator can be used to generate the object mask proposals. 

\subsection{Mining Object Mask Proposals}
\label{subsec: saliency}
We need to retrieve a set of object mask proposals for the images in our dataset. The literature~\cite{arbelaez2010contour,nguyen2019deepusps,pinheiro2015learning,van2011segmentation} offers a multitude of ways to do this. We prefer to use a simple strategy to verify whether unsupervised semantic segmentation benefits from adopting a mid-level visual prior. Moreover, we would like to use a method that does not rely on external supervision, or can be trained with a limited amount of annotations. In the latter case, the object mask proposal mechanism should generalize well to new scenes.

Based upon our requirements, we propose the use of saliency estimation~\cite{borji2015salient,wang2019salient} to generate object masks proposals. Most importantly, various unsupervised methods can be used for this purpose. Several of these works~\cite{nguyen2019deepusps,zhang2017supervision,zhang2018deep} used predictions obtained with hand-crafted priors~\cite{jiang2013saliency,li2013saliency,zhu2014saliency,zou2015harf} as pseudo-labels to train a deep neural network. Others~\cite{yang2020time,yang2019unsupervised} relied on videos to learn a salient object detector. Furthermore, on a variety of datasets~\cite{MSRA10K,MSRAB,ECSSD} unsupervised saliency methods have shown to perform on par with their supervised counterparts~\cite{hou2017deeply,luo2017non,qin2019basnet,wang2017stagewise,zhang2020uc,zhang2017amulet}. Finally, the model predictions transfer well to novel unseen datasets as shown by~\cite{nguyen2019deepusps}. 

\begin{figure}
    \centering
    \includegraphics[width=\linewidth]{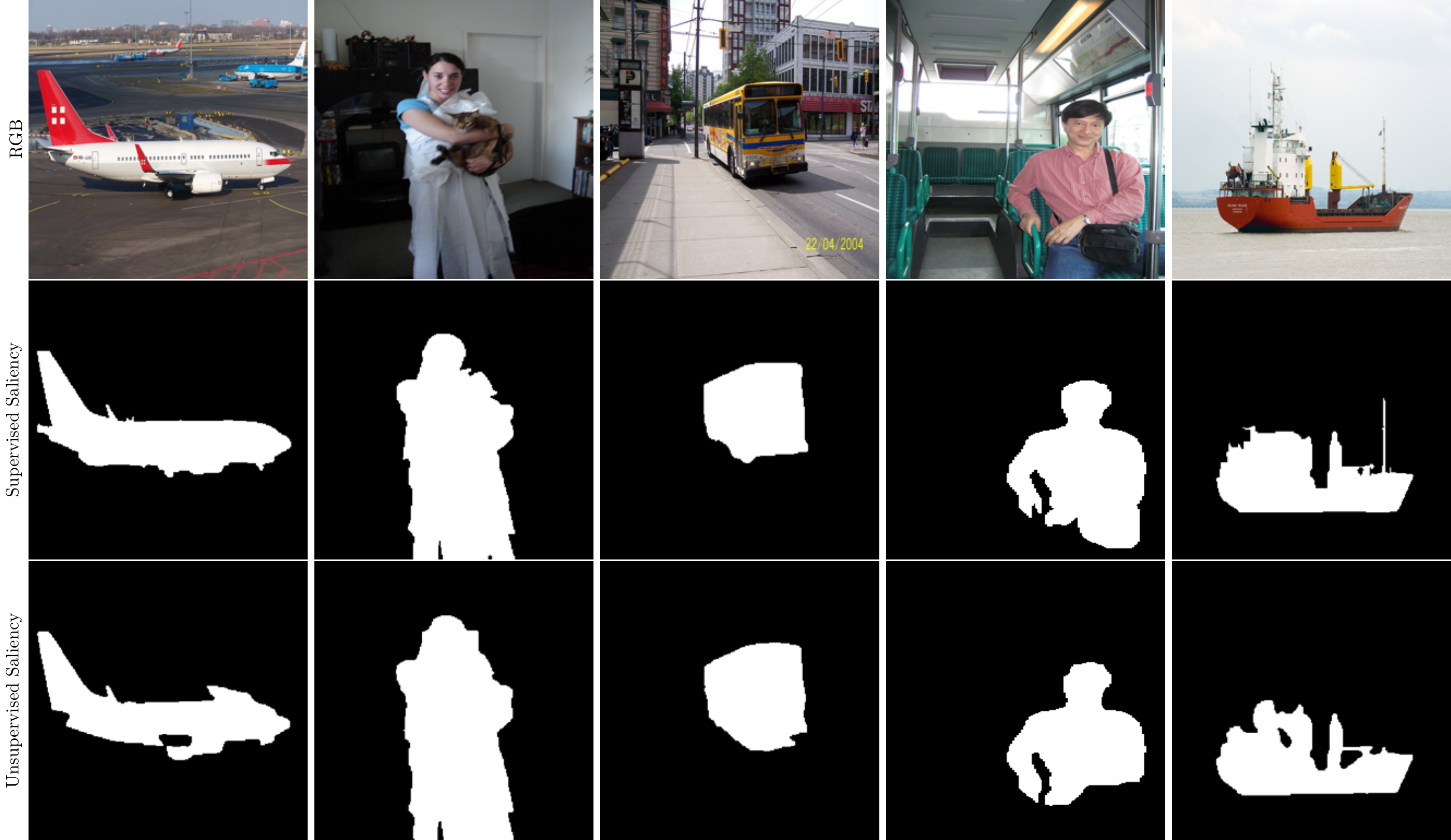}
    \caption{\textbf{Mask Proposals.} We train a supervised (middle) and unsupervised (bottom) saliency estimator on the DUTS and MSRA datasets respectively. We make predictions on PASCAL.}
    \label{fig: object_mask_proposals}
\end{figure}

For completeness, in Section~\ref{sec: experiments} we explore both unsupervised~\cite{nguyen2019deepusps} and supervised~\cite{qin2019basnet} saliency estimation methods to predict the object masks, and showcase the potential of our method. Figure~\ref{fig: object_mask_proposals} shows some examples. 

\subsection{MaskContrast: Learning Pixel Embeddings by Contrasting Salient Objects}
\label{subsec: self_supervised}
Consider a dataset of images $\mathcal{X}$ with associated non-overlapping object mask proposals $\left\{\mathcal{M}_0, \mathcal{M}_1, \ldots, \mathcal{M}_N \right\}$ obtained using a saliency estimator. Our goal is to learn a \emph{pixel embedding function} $\Phi_\theta: \mathcal{X} \to \mathcal{Z}$ parameterized by a neural network with weights $\theta$, that maps each pixel $i$ in an image to a point $\mathbf{z}_i$ on a $D$-dimensional normalized hyper-sphere. We chose a normalized embedding space, so that the output of $\Phi_\theta$ is bounded. Note that, the use of such scale-invariant embeddings decouples the loss from other design choices that could implicitly limit the range of distances, e.g. weight decay, as shown in~\cite{kong2018recurrent}. 

We construct an optimization objective to learn the embedding function $\Phi_\theta$ as follows. First, we describe how to learn semantically meaningful image feature using a contrastive learning objective. Second, we modify the criterion to learn pixel embeddings. 

\paragraph{Learning Image-Level Representations.} Existing contrastive self-supervised methods (e.g.~\cite{chen2020simple,he2019momentum,wu2018unsupervised}) learn visual representations through an instance discrimination task defined at the image-level. Positive views $(X, X^+)$ of the same image are acquired for which it is guaranteed that both images contain a part of the same object. Similarly, examples of negative pairs $\left\{(X, X^-_0), (X, X^-_1), \ldots, (X, X^-_K)\right\}$ can be found that never contain the same object. In practice, we impose additional invariances by applying augmentations. The positives and negatives can now be used in a contrastive framework to learn image representations that encode semantic information about the objects. 

We realize this concept by training an \emph{image embedding function} $\Psi_\eta$ to maximize the agreement between positive pairs $(X, X^+)$, while minimizing the agreement between negative pairs $\left\{(X, X^-_0), (X, X^-_1), \ldots, (X, X^-_K)\right\}$. If we measure the similarity between pairs using a dot product, the contrastive loss~\cite{gutmann2010noise,oord2018representation} is defined as 
\begin{equation}
\label{eq: contrastive_loss}
\mathcal{L} = - \log{\frac{\exp(\Psi_\eta(X)^T \cdot \Psi_\eta(X^+)/\tau)}{\sum_{k=0}^K \exp(\Psi_\eta(X)^T \cdot \Psi_\eta(X^-_k)/\tau)}}, 
\end{equation}
where the temperature $\tau$ relaxes the dot product. As shown by~\cite{purushwalkam2020demystifying}, the model learns to encode object information because the positive examples always preserve a part of the same object. Moreover, since the representational capacity of the network is intentionally limited, visually similar objects will tend to be mapped closer together by $\Psi_\eta$. The combination of these two properties results in image representations that can be directly clustered into semantic groups (see also~\cite{van2020scan} for a more detailed explanation). 

The above observations showed how to train a model that encodes semantic object information. Next, we modify the contrastive loss from Equation~\ref{eq: contrastive_loss} to learn representations at the pixel level.

\paragraph{Learning Pixel-Level Representations.} We adopt the following notation. Let $i$ be a pixel with $\mathbf{z_i}$ its pixel embedding. Let $m(i)$ be the index of the object mask that pixel $i$ belongs to, i.e. $i \in \mathcal{M}_{m(i)}$. Finally, let the mean pixel embedding $\mathbf{z}_{\mathcal{M}_n}$ of an object mask $\mathcal{M}_n$ be defined as
\begin{equation} 
\label{eq: mean_pixel_embedding}
\mathbf{z}_{\mathcal{M}_n} = \frac{1}{|\mathcal{M}_n|} \sum_{i \in \mathcal{M}_n} \mathbf{z}_i.
\end{equation} 

The optimization objective is derived from a pull- and push-force in the pixel embedding space.

\textbf{Pull-force.} In Section~\ref{subsec: prior}, we motivated the use of a prior based upon \emph{shared pixel ownership} to pull pixels together in the embedding space. More concretely, if two pixels $i, j$ belong to the same object, i.e. $m(i) = m(j)$, we maximize the agreement between their pixel embeddings $\mathbf{z}_i, \mathbf{z}_j$. In practice, the agreement is maximized between pixels and the mean embedding of their object mask in order to obtain a criterion that scales linearly with the number of pixels, rather than quadratically.

\textbf{Push-force.} Additionally, we require a push-force to avoid mode collapse in the embedding space. Moreover, the push-force should drive pixels from visually similar objects to lie close together in the embedding space, while pixels from dissimilar objects to be mapped further apart. As motivated in the previous paragraph, this can be achieved by adopting a contrastive loss that takes augmented views of objects as positive pairs, and views of other objects as negatives. In this case, the push-force is found between different objects. We represent the objects by their mean pixel embedding. 

\textbf{Optimization objective.} We modify the contrastive loss from Equation~\ref{eq: contrastive_loss} to include the proposed pull- and push-forces. Positive pairs of object-centric crops $(\Psi_\eta(X), \Psi_\eta(X^+))$ are replaced with positive pairs of pixel embeddings: $(\mathbf{z}_i, \mathbf{z}_{\mathcal{M}_{X^+}})$ for $i \in \mathcal{M}_X$. In a similar way, the negative pairs $(\Psi_\eta(X), \Psi_\eta(X^-_k))$ are replaced with $(\mathbf{z}_i, \mathbf{z}_{\mathcal{M}_{X^-_k}})$. We obtain the following optimization criterion for a pixel $i \in \mathcal{M}_X$
\begin{equation}
\label{eq: pixel_objective}
\mathcal{L}_{i} = -\log \frac{\exp \left( \mathbf{z}_i \cdot \mathbf{z}_{\mathcal{M}_{X^+}} /\tau \right)}{\sum_{k=0}^{K} \exp \left( \mathbf{z}_i \cdot \mathbf{z}_{\mathcal{M}_{X^-_k}} / \tau \right)}.
\end{equation}
The pixel embedding function $\Phi_\theta$ maximizes the agreement between pixels and an augmented view of the object they belong to, while minimizing the agreement with other objects. We apply the pixel-wise loss $\mathcal{L}_i$ to all foreground pixels. The background pixels are not contrasted, since there could be multiple background objects on which we have no conclusive information. In this case, however, the network does not need to discriminate between pixels that fall inside or outside the object masks. As a consequence, the pixel embeddings can collapse to a single vector across an image. To prevent this, we regularize the feature space by including a separate linear head that predicts the saliency masks. We refer to the pseudo-code in the supplementary materials for an overview of MaskContrast (Algorithm~\ref{alg:code}).

Interestingly, the proposed objective can also be viewed in an alternative way. Wang and Isola~\cite{wang2020understanding} showed that a contrastive loss optimizes two properties: (1) alignment of features from positive pairs and (2) uniformity of the feature distribution on a normalized hyper-sphere. From this viewpoint, our optimization objective can also be interpreted as optimizing the alignment of pixel embeddings based upon shared pixel ownership, while spreading pixel embeddings uniformly across the hyper-sphere $\mathcal{Z}$.

%% file: sections/experiments.tex
\section{Experiments}
\label{sec: experiments}

\subsection{Experimental Setup}
\label{subsec: experiments_setup}

\paragraph{Datasets.} We conduct the bulk of our experimental analysis on the PASCAL~\cite{everingham2010pascal} dataset following prior work~\cite{hwang2019segsort,zhang2020self}. The \texttt{train\_aug} and \texttt{val} splits are used during training and evaluation respectively. We perform additional experiments on the COCO~\cite{lin2014microsoft} and DAVIS-2016~\cite{perazzi2016} datasets to verify if the pixel embeddings transfer to novel scenes. We use the annotations from Kirillov~\etal~\cite{kirillov2019panoptic} for the semantic segmentation task on COCO and evaluate on the PASCAL classes. On DAVIS-2016, the representations are used to compute correspondences for propagating object masks in videos. Only the first frame is annotated and we evaluate the propagated masks on the rest of the frames. We adopt the evaluation protocol from~\cite{jabri2020walk}, and report the region similarity $\mathcal{J}$ and contour-based accuracy $\mathcal{F}$ scores.

\paragraph{Training setup.} We use a DeepLab-v3~\cite{chen2017rethinking} model with dilated~\cite{yu2015multi} ResNet-50 backbone~\cite{he2016deep}. The backbone is initialized from MoCo v2~\cite{chen2020improved} pre-trained on ImageNet, unless defined otherwise. We train the model for $60$ epochs using batches of size $64$. The model weights are updated through SGD with momentum $0.9$ and weight decay $1e^{-4}$. The initial learning is set to $0.004$ and decayed with a poly learning rate scheme. We use the same set of augmentations as SimCLR~\cite{chen2020simple} to generate positive pairs $(X, X^+)$, while making sure that each image contains at least a part of the salient object $(\text{area} > 10\%)$. The features of negatives $\left\{\mathbf{z}_{\mathcal{M}_{X^-_0}}, \ldots, \mathbf{z}_{\mathcal{M}_{X^-_K}} \right\}$ are saved in a memory bank, with $K$ set to $128$. The negatives are encoded with a momentum-updated version of the network following~\cite{he2019momentum}. We use dimension $D=32$ and temperature $\tau=0.5$.

\paragraph{Saliency estimation.} We test both unsupervised and supervised saliency estimators to mine the object mask proposals. We adopt the BAS-Net~\cite{qin2019basnet} architecture. The \emph{supervised saliency model} is trained on DUTS~\cite{wang2017learning}. Differently, the \emph{unsupervised saliency model} is trained on MSRA~\cite{MSRA10K} using the approach from DeepUSPS~\cite{nguyen2019deepusps}. MSRA considers less complex scenes from which the unsupervised training benefits. However, directly transferring the predictions to our target datasets, e.g. PASCAL, results in low-quality mask proposals when using the unsupervised model. We employ a simple bootstrapping procedure to improve the predictions on the target datasets. In particular, we obtain our final saliency estimator from training BAS-Net on pseudo-labels generated with the unsupervised DeepUSPS model on MSRA. 

\paragraph{Implementation.} We provide the implementation details of every method in the supplementary materials. The code and pre-computed saliency masks will be made available.

\paragraph{Scope.} We adopt standard evaluation protocols~\cite{xu2019invariant,zhang2020self} for unsupervised semantic segmentation to benchmark our method. More specifically, we use linear probes (Sec.~\ref{subsec: experiments_linear}), direct clustering (Sec.~\ref{subsec: experiments_clustering}) and a segment retrieval approach (Sec.~\ref{subsec: experiments_retrieval}) to quantify if the pixel embeddings are disentangled according to the semantic classes. This experimental setup differs from the typical setting used in self-supervised representation learning, where the evaluation focuses on fine-tuning the feature representations to various down-stream tasks. For completeness, we include additional fine-tuning experiments in Sections~\ref{subsec: experiments_transfer}~-~\ref{subsec: experiments_semi_sup}.

\subsection{Ablation Studies}
\label{subsec: experiments_ablations}
We examine the influence of the different components of our framework under the linear evaluation protocol following existing work~\cite{zhang2020self}. The network weights are kept fixed and we train a 1 x 1 convolutional layer on top to predict the class assignments. Since the discriminative power of a linear classifier is low, the pixel embeddings need to be informative of the semantic class to solve the task in this way. 

\newlength\savewidth\newcommand\shline{\noalign{\global\savewidth\arrayrulewidth
\global\arrayrulewidth 1pt}\hline\noalign{\global\arrayrulewidth\savewidth}}
\newcommand{\tablestyle}[2]{\setlength{\tabcolsep}{#1}\renewcommand{\arraystretch}{#2}\centering\footnotesize}

\begin{table}
    \tablestyle{4pt}{1.0}
    \begin{tabular}{l|ccc}
    \textbf{Method}  & \textbf{LC (MIoU)} \\
    \shline
    Supervised Saliency Model & 6.5 \\
    \hline
    MoCo v2~\cite{chen2020improved} (Unsupervised) & 45.0 \\
    ImageNet (IN) Classifier (Supervised) & 53.1 \\
    \hline
    MaskContrast (MoCo v2 Init. - Unsup. Sal. Model) & 58.4 \\
    MaskContrast (MoCo v2 Init. - Sup. Sal. Model) & 62.2 \\
    \hline
    MaskContrast (IN Classifier Init. - Unsup. Sal. Model) & 61.0 \\
    MaskContrast (IN Classifier Init. - Sup. Sal. Model) & \textbf{63.9} \\
    \end{tabular}
    \vspace{0.3em}
    \caption{\textbf{Baseline comparison} under the linear evaluation protocol on PASCAL.}
    \label{tab: ablation_baselines}
\end{table}

\paragraph{Baseline comparison.} Table~\ref{tab: ablation_baselines} compares several baselines. Applying a linear classifier on top of the saliency features results in the lowest performance ($6.5\%$). This is to be expected since the saliency estimator only discriminates between two groups of pixels, i.e. the salient object vs. background. Differently, our method discovers a semantically structured embedding space, where pixels from visually similar objects lie close together, while pixels from dissimilar objects end up far apart. This allows a linear classifier to correctly group the pixels ($> 58.4 \%$). Importantly, the results improve over the models from which the backbone weights were initialized ($45.0\%$ to $58.4\%$ for MoCo and $53.1\%$ to $61.0\%$ for supervised pre-training). We conclude that the performance of our method can not be attributed to the use of a specific initialization. Also, it is beneficial to learn representations at pixel-, rather than at image-level, for the segmentation task. Finally, we observe further performance gains when including additional supervision, e.g. supervised pre-training on ImageNet ($58.4\%$ to $61.0\%$), or a supervised saliency estimator ($58.4\%$ to $62.2\%$ and $61.0\%$ to $63.9\%$).

\begin{table*}[t]
\vspace{-0.3em}
\resizebox{1.0\linewidth}{!}{
\subfloat[\centering Comparison of three mask proposal mechanisms.]{\adjustbox{raise=.5em}{
\tablestyle{4pt}{1.0}
\label{subtab: mask_proposals}
\begin{tabular}{l|ccc}
\textbf{Mask Proposals} & \textbf{LC} \\
& \textbf{(MIoU)} \\
\shline
Hierarchical Seg.~\cite{arbelaez2010contour,xie2015holistically} & 30.5 \\
Unsupervised Sal. Model & 58.4 \\
Supervised Sal. Model & 62.2 \\
\end{tabular}
}}
\subfloat[\centering Analysis of the used training mechanisms.]{
\tablestyle{4pt}{1.0}
\label{subtab: train_mechanisms}
\begin{tabular}{c|c|c|c}
\textbf{Augmented} & \textbf{Memory} &\textbf{Momentum} & \textbf{LC} \\
\textbf{Views} & & \textbf{Encoder} & \textbf{(MIoU)} \\
\shline
\xmark & \xmark & \xmark & 52.4 \\
\checkmark & \xmark & \xmark & 54.0 \\
\checkmark & \checkmark & \xmark & 55.0 \\
\checkmark & \checkmark & \checkmark & 58.4 \\
\end{tabular}
}
\subfloat[\centering Hyperparameter study. We report the mean and standard deviation.]{\adjustbox{raise=.93em}{
\tablestyle{4pt}{1.0}
\begin{tabular}{l|l|c}
\label{subtab: hyperparams}
\textbf{Hyperparameter} & \textbf{Range} & \textbf{LC} \\ &&\textbf{(MIoU)} \\
\shline
Temperature $\tau$ & [0.1-1] & $56.2 \pm 1.4$\\
Negatives $K$ & [64-1024] & $57.0\pm 0.6$  \\
\end{tabular}
}}
}
\vspace{0.3em}
\caption{\textbf{Ablation studies} of our method under the linear evaluation protocol on PASCAL. Tables~\ref{subtab: train_mechanisms}-~\ref{subtab: hyperparams} report results with masks from the unsupervised saliency estimator. We use MoCo~v2 initial weights.}
\label{tab: ablation_method}
\end{table*}

\paragraph{Mask proposals.} Table~\ref{subtab: mask_proposals} compares three mask proposal strategies. Better numbers are reported when using salient object masks. We found that the regions extracted with the hierarchical segmentation algorithm were often too small to be representative of an object or part. In this way, the model does not learn useful information for the segmentation task. This confirms the hypothesis from Section~\ref{subsec: prior}, i.e. a good prior expresses object information.

\paragraph{Training mechanisms.} Table~\ref{subtab: train_mechanisms} ablates some of the included training mechanisms. First, using augmented views to sample positive pairs improves the results, as we learn additional invariances. Second, including a memory bank results in further performance gains, because we can better estimate the distribution of negatives. Third, it is helpful to encode the negatives with a momentum-updated version of the network $\Phi_\theta$, as this enforces consistency in the memory bank (see also~\cite{he2019momentum}). In summary, all three mechanisms positively contribute to the results. 

\paragraph{Hyperparameter study.} Table~\ref{subtab: hyperparams} studies the influence of the used temperature $\tau$ and number of negatives $K$. We conclude that the proposed algorithm is not very hyperparameter sensitive based upon the reported standard deviations.

\subsection{Linear Classifier}
\label{subsec: experiments_linear}
Table~\ref{tab: linear_probe} compares our method against competing works under the linear evaluation protocol on PASCAL. 

\textbf{MaskContrast vs.~proxy tasks.} The method substantially outperforms works based on proxy tasks. It is unlikely that a proxy task aligns the embeddings with the semantic groups in the dataset. In contrast, combining our proposed prior, i.e. shared pixel ownership, with a contrastive loss results in more semantically meaningful pixel embeddings.

\textbf{MaskContrast vs.~clustering.} We outperform IIC~\cite{xu2019invariant} which used a clustering objective. As discussed earlier, the clusters strongly depend on the network initialization, which negatively impacts the learned features as the network can latch onto low-level information, like color, texture, contrast, etc. Differently, we suppress these problems by decoupling the prior from the network initialization. 

\textbf{MaskContrast vs.~contrastive learning.} The method reports higher accuracy compared to existing contrastive self-supervised approaches. This group of works defined the contrastive loss at the global image- or patch-level. Naturally, our pixel embeddings are more predictive of the semantic segmentation task as we defined a contrastive learning objective at the pixel-level. 

\textbf{MaskContrast vs.~boundary based.} Finally, we outperform methods that relied on boundary detectors to group pixels together. We argue that the employed saliency masks incorporate higher level visual information compared to the regions obtained from boundary detectors. The results support our earlier claims from Section~\ref{subsec: prior}.

\subsection{Clustering}
\label{subsec: experiments_clustering}
We verify whether the feature representations can be directly clustered in semantically meaningful groups using an off-line clustering criterion like K-Means. The number of clusters equals the number of ground-truth classes. The Hungarian matching algorithm is used to match the predicted clusters with the ground-truth classes and the results are averaged across five runs. Table~\ref{subtab: clustering} shows the results. Our learned pixel embeddings can be successfully clustered using K-Means on PASCAL. In contrast, the features representations obtained in prior works do not exhibit this behavior. We include additional results in the suppl. materials when applying overclustering. 

\begin{table}
\vspace{-1.5em}
\hspace{-0.6em}
\resizebox{1.0\linewidth}{!}{
\subfloat[\centering Linear classifier.]{
    \tablestyle{4pt}{1.0}
    \label{tab: linear_probe}
    \begin{tabular}{l|l}
    \textbf{Method} & \textbf{LC} \\
    \shline
    \multicolumn{1}{l}{\emph{Proxy task based:}}\\
    \hline
    Co-Occurence~\cite{isola2015learning} & 13.5 \\
    CMP~\cite{zhan2019self} & 16.5 \\
    Colorization~\cite{zhang2016colorful} & 25.5 \\
    \hline
    \multicolumn{2}{l}{\emph{Clustering based:}} \\
    \hline
    IIC~\cite{xu2019invariant} & 28.0 \\
    \hline
    \multicolumn{2}{l}{\emph{Contrastive learning based:}} \\
    \hline
    Inst. Discr.~\cite{wu2018unsupervised} & 26.8 \\
    MoCo v2~\cite{he2019momentum} & 45.0 \\
    InfoMin~\cite{tian2020infomin} & 45.2 \\
    SWAV~\cite{caron2020unsupervised} & 50.7 \\
    \hline
    \multicolumn{2}{l}{\emph{Boundary based:}}\\
    \hline
    SegSort~\cite{hwang2019segsort}$^\dagger$ & 36.2 \\
    Hierarch. Group.~\cite{zhang2020self}$^\dagger$ & 48.8 \\
    \hline
    ImageNet (IN) Classifier (Supervised) & 53.1 \\
    \hline
    MaskContrast (MoCo Init. + Unsup. Sal.) & 58.4 \\
    MaskContrast (MoCo Init. + Sup. Sal.) & 62.2 \\
    MaskContrast (IN Sup. Init. + Unsup. Sal.) & 61.0 \\
    MaskContrast (IN Sup. Init. + Sup. Sal.) & \textbf{63.9} 
    \end{tabular}}

\subfloat[\centering K-Means.]{
    \tablestyle{4pt}{1.0}
    \begin{tabular}{c}
    \textbf{K-Means} \\
    \shline
    \\
    \hline
    4.0 \\
    4.3 \\
    4.9 \\
    \hline
    \\
    \hline
    9.8 \\
    \hline
    \\
    \hline
    4.4 \\
    4.3 \\
    3.7 \\
    4.4 \\
    \hline
    \\
    \hline
    - \\
    - \\
    \hline
    4.7 \\
    \hline
    35.0 \\
    38.9 \\
    41.6 \\
    \textbf{44.2}\\
    \end{tabular}
    \label{subtab: clustering}
}}
\vspace{0.3em}
\caption{\textbf{State-of-the-art comparison} on PASCAL \emph{val} (MIoU). ($\dagger$) Indicates results taken from~\cite{zhang2020self}. Note that the authors use a slightly different evaluation protocol, i.e. without ImageNet pretraining, but with finetuning of the complete ASPP decoder.}
\end{table}

\subsection{Semantic Segment Retrieval}
\label{subsec: experiments_retrieval}
\begin{figure}
\centering
\includegraphics[width=0.9\linewidth]{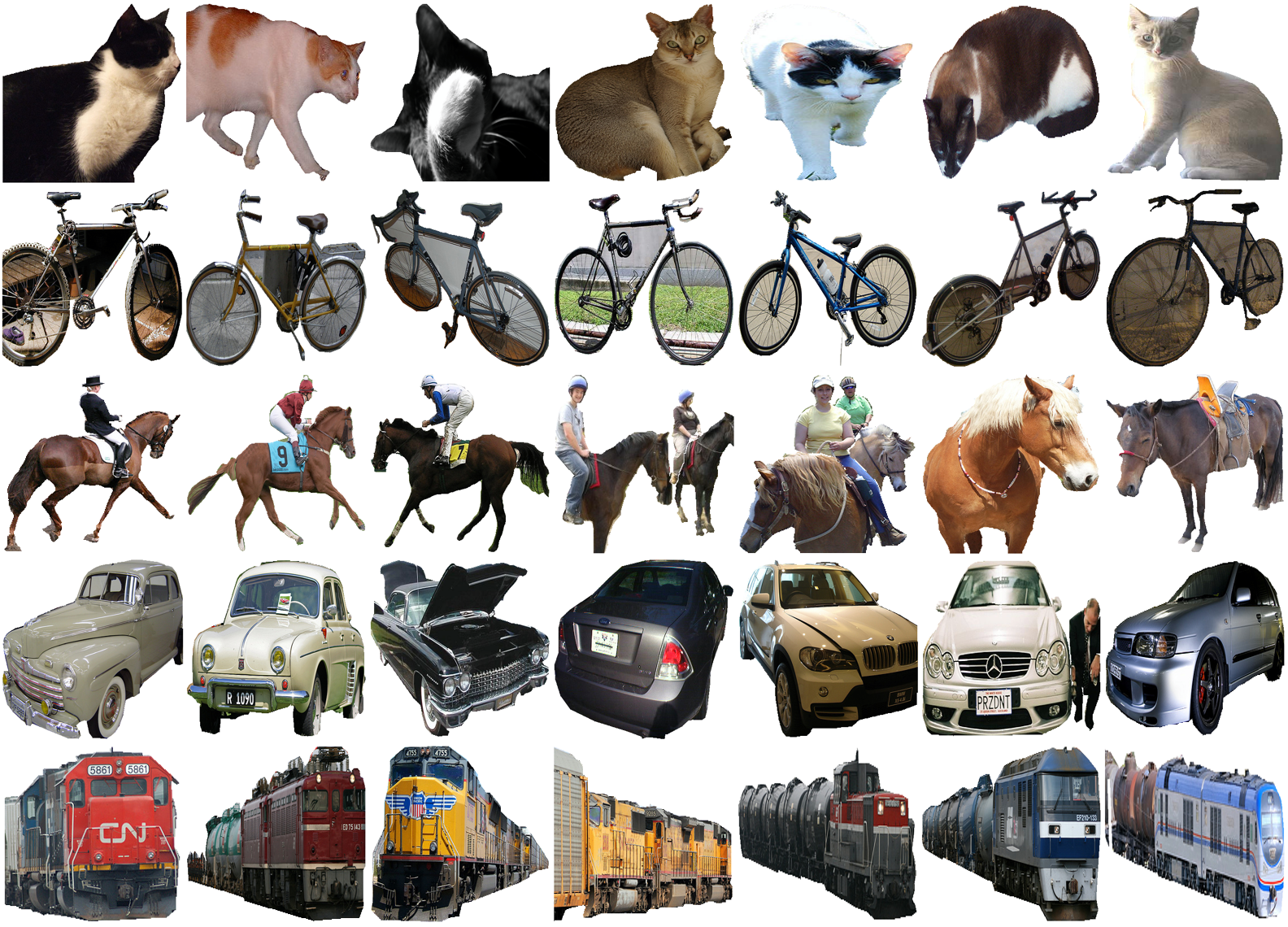}
\vspace{0.3em}
\caption{\textbf{Nearest neighbors} for queries (1st col.) on PASCAL.}
\label{fig: nearest_neighbors}
\end{figure}

Next, we adopt a retrieval approach to examine our representations on PASCAL. First, we compute a feature vector for every salient object by averaging the pixel embeddings within the predicted mask. Next, we retrieve the nearest neighbors of the \texttt{val} set objects from the \texttt{train\_aug} set. Table~\ref{tab: retrieval} shows a quantitative comparison with the state-of-the-art for the following 7 classes: bus, airplane, car, person, cat, cow and bottle. As before, we outperform prior works by significant margins. To facilitate future comparison, we also include results when evaluating on all 21 PASCAL classes. Figure~\ref{fig: nearest_neighbors} shows some qualitative results. 

\begin{table}
\resizebox{1.0\linewidth}{!}{
\tablestyle{4pt}{1.0}
\centering
\begin{tabular}{l|c|c}
\textbf{Method} & \textbf{MIoU (7 classes)} & \textbf{MIoU (21 classes)} \\
\shline
SegSort~\cite{hwang2019segsort} & 10.2 & - \\
Hierarch. Group.~\cite{zhang2020self} & 24.6 & - \\
MoCo v2~\cite{chen2020improved} & 48.0 & 39.0 \\
\hline
MaskContrast (Unsup. Sal.) & 53.4 & 43.3 \\
MaskContrast (Sup. Sal. ) & \textbf{62.3} & \textbf{49.6} \\
\end{tabular}}
\vspace{0.3em}
\caption{\textbf{State-of-the-art comparison} for semantic segment retrieval on the PASCAL \texttt{val} set. We use MoCo v2 initial weights.}
\label{tab: retrieval}
\end{table}

\subsection{Transfer Learning}
\label{subsec: experiments_transfer}
We study the transferability of our pixel embeddings. Table~\ref{tab: transfer_learning} shows the results when pretraining on ImageNet and evaluating the generated pixel embeddings on a different target dataset. Interestingly, our representations transfer well across various datasets. Training a linear classifier to solve the segmentation task on PASCAL improves over the MoCo v2 baseline (55.4\% for MaskContrast vs. 45.0\% for MoCo when using an unsupervised saliency model). A similar effect can be observed on COCO (45.0\% for MaskContrast vs. 35.2\% for MoCo). Finally, our representations also transfer well to the semantic object segmentation task on DAVIS-2016. This dataset covers a rich set of natural image augmentations like viewpoint changes, occlusions, etc., for which our pixel embeddings have learned invariances. 

The gains observed across all three benchmarks show that the learned representations are not limited to a specific dataset. We conclude that the use of a mid-level visual prior can be useful for self-supervised representation learning. 

\begin{table}
    \tablestyle{4pt}{1.0}
    \centering
    \begin{tabular}{l|c|c|cc}
    \textbf{Model} & \textbf{PASCAL} & \textbf{COCO} & \multicolumn{2}{c}{\textbf{DAVIS '16}} \\
          & \textbf{(MIoU)}$\uparrow$ & \textbf{(MIoU)}$\uparrow$ & $\mathbf{\mathcal{J}_{m}}\uparrow$ & $\mathbf{\mathcal{F}_{m} \uparrow}$\\
    \shline
    MoCo v2 & 45.0 & 35.2 & 77.1 & 77.2 \\
    \hline
    MaskContrast (Unsup. Sal.) & 55.4 & 45.0 & 78.0 & 77.8 \\
    MaskContrast (Sup. Sal.) & \textbf{57.2} & \textbf{47.}2 & \textbf{82.0} & \textbf{80.9} \\
    \end{tabular}
    \vspace{0.3em}
    \caption{\textbf{Transfer learning setup.} All models were pre-trained on ImageNet. We use MoCo v2 initial weights. Results on PASCAL and COCO are reported for a linear classifier. On DAVIS, we freeze the representations and adopt the protocol from~\cite{jabri2020walk}.}
    \label{tab: transfer_learning}
\end{table}

\subsection{Semi-Supervised Learning}
\label{subsec: experiments_semi_sup}
The proposed method can alternatively be used as a pre-training strategy for semantic segmentation. That is, the model is fine-tuned in a semi-supervised way on PASCAL. We use 1\%, 2\%, 5\%, 12.5\% and 100\% of the \texttt{train\_aug} split as labeled examples. We initialize our model from supervised pre-training on ImageNet. This weight initialization is commonly used in semantic segmentation. Furthermore, directly fine-tuning a model initialized in the same way serves as a strong baseline. Table~\ref{tab: semi_supervised} shows the results.

The representations generated with our method yield higher performance after fine-tuning, compared to supervised pre-training on ImageNet. This holds true when using both an unsupervised and supervised saliency estimator to predict the object mask proposals. Predictably, the gains become smaller when more labeled examples are available (see also~\cite{zoph2020rethinking}). We conclude that unsupervised learning of pixel embeddings can complement an existing pre-training strategy based on an optimization criterion defined at the global image- or patch-level. We hope that this observation can spur further research efforts in this direction.

\begin{figure}
    \centering
    \includegraphics[width=\linewidth]{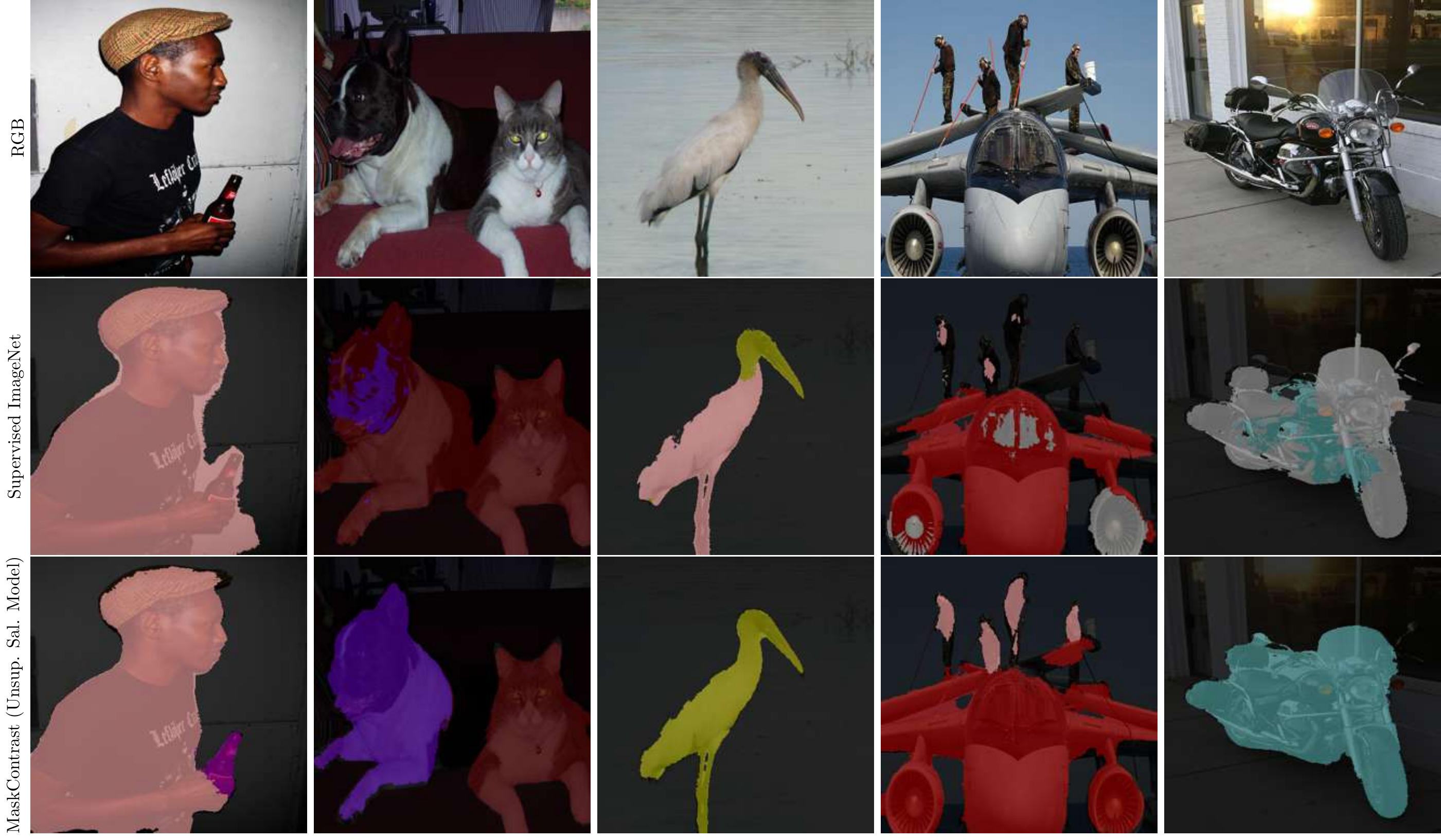}
    \caption{\textbf{Qualitative comparison} after fine-tuning on PASCAL using 1 \% of labeled data. We use supervised pre-training on ImageNet (middle) or our method (bottom) to initialize the weights before fine-tuning.}
    \label{fig: segmentation}
\end{figure}

\begin{table}
    \tablestyle{4pt}{1.0}
    \begin{tabular}{l|lllll}
    \textbf{Label Fraction} & 
    \textbf{1\%} & 
    \textbf{2\%} & 
    \textbf{5\%} &
    \textbf{12.5\%} &
    \textbf{100\%} \\
    \shline
    ImageNet Classifier Init. & 43.4  & 55.2 & 62.7 & 68.4 & 78.0 \\
    + MaskContrast (Unsup. Sal.)  & 50.5 & 57.2 & 64.5 & 69.0 & 78.4 \\
    + MaskContrast (Sup. Sal.) & \textbf{51.5} & \textbf{59.6} & \textbf{65.3} &  \textbf{69.4} & \textbf{78.6} \\
    \end{tabular}
    \vspace{0.3em}
    \caption{\textbf{Semi-supervised fine-tuning} on PASCAL (MIoU).}
    \label{tab: semi_supervised}
\end{table}

%% file: sections/conclusion.tex
\section{Discussion and Limitations}
\label{sec: conclusion}
This work presented a general two-step framework based upon a mid-level visual prior for tackling unsupervised semantic segmentation. The proposed setup prevents the model from latching onto low-level image features, a problem present in prior works that relied on end-to-end clustering, proxy tasks or low-level visual cues. Instead, MaskContrast learns pixel embeddings which incorporate more semantically meaningful information (see Figure~\ref{fig: nearest_neighbors}). As a result, we were able to tackle the semantic segmentation task under a fully unsupervised setup on a diverse dataset like PASCAL. Further, experimental evaluation showed that our pixel embeddings have several other interesting properties: the ability for semantic segment retrieval, transfer learning and semi-supervised fine-tuning. 

Still, there are some limitations to our method. The object mask proposals were obtained using a salient object estimator - which can retrieve only a limited number of objects per image.  Alternative ways to mine the object mask proposals can be explored for tackling even more challenging datasets where many objects can exist per image. In particular, we could see additional sensory data~\cite{tian2020unsupervised} or other techniques~\cite{pinheiro2015learning} being used that are better suited for this type of images. The optimization criterion from Equation~\ref{eq: pixel_objective} could then be extended accordingly. Given the viability of our framework, we believe these are interesting research directions. 

%% file: supp.tex
The supplementary materials include the pseudocode of our algorithm and the implementation details of the experimental evaluation. Additionally, we provide overclustering results in Section~\ref{sec: clustering} and qualitative results for semi-supervised fine-tuning in Section~\ref{sec: qualitative_results}.

\begin{algorithm}[t]
\caption{Pseudocode of~\textbf{MaskContrast}.}
\label{alg:code}
\algcomment{\fontsize{7.2pt}{0em}\selectfont \texttt{bmm}: batch matrix multiplication; \texttt{mm}: matrix multiplication; \texttt{cat}: concatenation; \texttt{BCE}: binary cross-entropy loss; \texttt{remap}: custom function. 
}
\definecolor{codeblue}{rgb}{0.25,0.5,0.5}
\lstset{
  backgroundcolor=\color{white},
  basicstyle=\fontsize{7.2pt}{7.2pt}\ttfamily\selectfont,
  columns=fullflexible,
  breaklines=true,
  captionpos=b,
  commentstyle=\fontsize{7.2pt}{7.2pt}\color{codeblue},
  keywordstyle=\fontsize{7.2pt}{7.2pt},
}
\begin{lstlisting}[language=python]
# f_q, f_k: encoder-decoder networks for query and key
# queue: dictionary of K prototype keys (CxK)
# m: momentum
# t: temperature
# H, W = height, width of an image x
# P : number of salient pixels in a batch

f_k.params = f_q.params  # initialize
for (x, s) in loader:  
    # load a batch with N samples and N saliency masks
    # constrain aug s.t. object area > threshold
    x_q, s_q = aug(x, s)  # augmented version
    x_k, s_k = aug(x, s)  # another augmented version
    
    q, aux = f_q.forward(x_q)  # q: NxCxHxW, aux: NxHxW
    k, _ = f_k.forward(x_k) # k: NxCxHxW
    
    # salient objects are non-zero
    valid_ids = s_q.nonzero() # valid_ids: Px1
    # remap each object to a unique id in {0..N-1}
    s_r = remap(s_q) # s_r: Px1
    
    # key prototypes: NxC
    p_k = bmm(k.view(N,C,H.W), s_k.view(N,H.W,1))
    p_k = normalize(p_k, dim=1) # L2-normalize
    p_k = p_k.detach()  # no gradient to prototypes
    
    # select embeddings of salient objects: PxC
    q = index_select(q.view(H.WxC), index=valid_ids)
    
    # positive logits: PxN
    l_pos = mm(q.view(P,C), p_k.view(C,N))

    # negative logits: PxK
    l_neg = mm(q.view(P,C), queue.view(C,K))

    # logits: Px(N+K)
    logits = cat([l_pos, l_neg], dim=1)

    # contrastive loss: positives are the s_r-th
    MaskContrast_loss = CrossEntropyLoss(logits/t, s_r) 
    
    # auxiliary BCE loss to prevent collapse
    aux_loss = BCE(aux, s_q)
    total_loss = MaskContrast_loss + aux_loss

    # SGD update: query network
    total_loss.backward()
    update(f_q.params)

    # momentum update: key network
    f_k.params = m*f_k.params+(1-m)*f_q.params

    # update dictionary
    enqueue(queue, p_k)  # enqueue current prototypes
    dequeue(queue)  # dequeue earliest prototypes

\end{lstlisting}
\end{algorithm}

\section{Pseudo-code}
Algorithm~\ref{alg:code} shows the pseudo-code of MaskContrast. The saliency masks are obtained by running the public code of existing saliency estimators~\cite{nguyen2019deepusps,qin2019basnet}. The default hyperparameter settings were used. A PyTorch implementation of our method and pre-computed saliency masks will be made publicly available. 

\section{Pre-training}
\label{sec: pretrain}
This section describes the pre-training setup for the models included in the experiments section of the main paper. In the majority of cases, we were able to use the pre-trained weights made available by the authors of the respective works. 
\vspace{1em}

\noindent\textbf{Co-Occurence.} We adopt the training setup from the original work~\cite{isola2015learning}. The features before the output layer of the network are used for the purpose of training a linear classifier and applying K-Means clustering. 

\vspace{1em}

\noindent\textbf{Colorization.} The pre-trained colorizer from Zhang~\etal~\cite{zhang2016colorful} is used. It is argued that the intermediate representations in the network will extract semantic information in order to solve the colorization task. As a consequence, it is non-trivial from what layer we should tap the features to tackle the semantic segmentation task. To resolve this, we tried using features from various intermediate layers, and report the best results when training a linear classifier or applying K-Means. 

\vspace{1em}

\noindent\textbf{CMP.} We follow the strategy from the colorization task for training a linear classifier or applying K-Means. The pre-trained model from Zhan~\etal~\cite{zhan2019self} is used.

\vspace{1em}

\noindent\textbf{IIC.} We follow the implementation strategy from~\cite{xu2019invariant}. 

\vspace{1em}

\noindent\textbf{Contrastive-Learning Methods.} We used the weights from a ResNet-50 model pre-trained on ImageNet. The weights were made available by the authors of the respective works, i.e. the instance discrimination task~\cite{wu2018unsupervised}, SWAV~\cite{caron2020unsupervised}, MoCo v2~\cite{he2019momentum} and InfoMin~\cite{tian2020infomin}. In some cases, multiple variants of the model were released, e.g. when using different augmentation strategies during training. We chose the best available model each time. 

The contrastive learning models were only pre-trained on ImageNet, as we could not see any substantial improvements from further pre-training them on the target dataset, i.e. PASCAL. To obtain dense predictions, we apply dilated convolutions in the last residual block. We use the features from the backbone for training a linear classifier or applying K-Means. 

\vspace{1em}

\noindent\textbf{MaskContrast.} We use a dilated ResNet-50 model with DeepLab-v3 head as outlined in the main paper. The final 1 x 1 convolutional layer is split into two linear heads. The first head predicts the pixel embeddings, while the second head predicts the saliency mask. During linear evaluation, we replace the final layer by a randomly initialized 1 x 1 convolutional layer. Other details were already provided in the paper. 

\section{Linear Classifier}
\label{sec: linear}
This section describes the training setup used for the linear evaluation protocol. We train a 1 x 1 convolutional layer for 60 epochs using batches of size 16. The complete train set is used during training. We optimize the weights through stochastic gradient descent with momentum $0.9$, weight decay $0.0001$ and initial learning rate $0.1$. The learning rate is reduced to $0.01$ after 40 epochs. We found that increasing the train time, or modifying the learning rate did not improve the results. 

\section{Clustering}
\label{sec: clustering}
This section specifies how to obtain discrete class assignments by clustering the representations using K-Means. We follow the evaluation strategy from~\cite{xu2019invariant} to calculate the mean IoU metric. In particular, we first match the predicted clusters with the ground-truth classes using a Hungarian algorithm. We subsequently calculate the mean IoU from the re-assigned clusters and the ground-truth labels. We report the average from five runs. 

\vspace{1em}
\noindent \textbf{Contrastive based methods.} As described in Section~\ref{sec: pretrain}, we apply K-Means clustering to the backbone features. The cluster assignments are upsampled to match the original image resolution, before applying the Hungarian algorithm. 

\vspace{1em}

\noindent \textbf{IIC.} No specific post-processing is required. We simply match the predicted clusters with the ground-truth classes following the original work~\cite{xu2019invariant}.

\vspace{1em}

\noindent \textbf{Proxy-task based methods (Co-Occurence, Colorization, CMP).} We select the features for applying K-Means as described in Section~\ref{sec: pretrain}. The predictions are up-sampled to match the original image resolution before applying the Hungarian algorithm. 

\vspace{1em}
\noindent \textbf{MaskContrast.} We compute the mean embeddings of the foreground objects and apply K-Means using the L2-normalized feature vectors. All pixels belonging to the object are assigned the same label as the mean-pixel embedding after clustering. The predictions from the saliency estimation head are used to identify the background class. We match the predictions with the ground-truth classes using the Hungarian algorithm. 
\begin{figure*}[t]
    \centering
    \includegraphics[width=\linewidth]{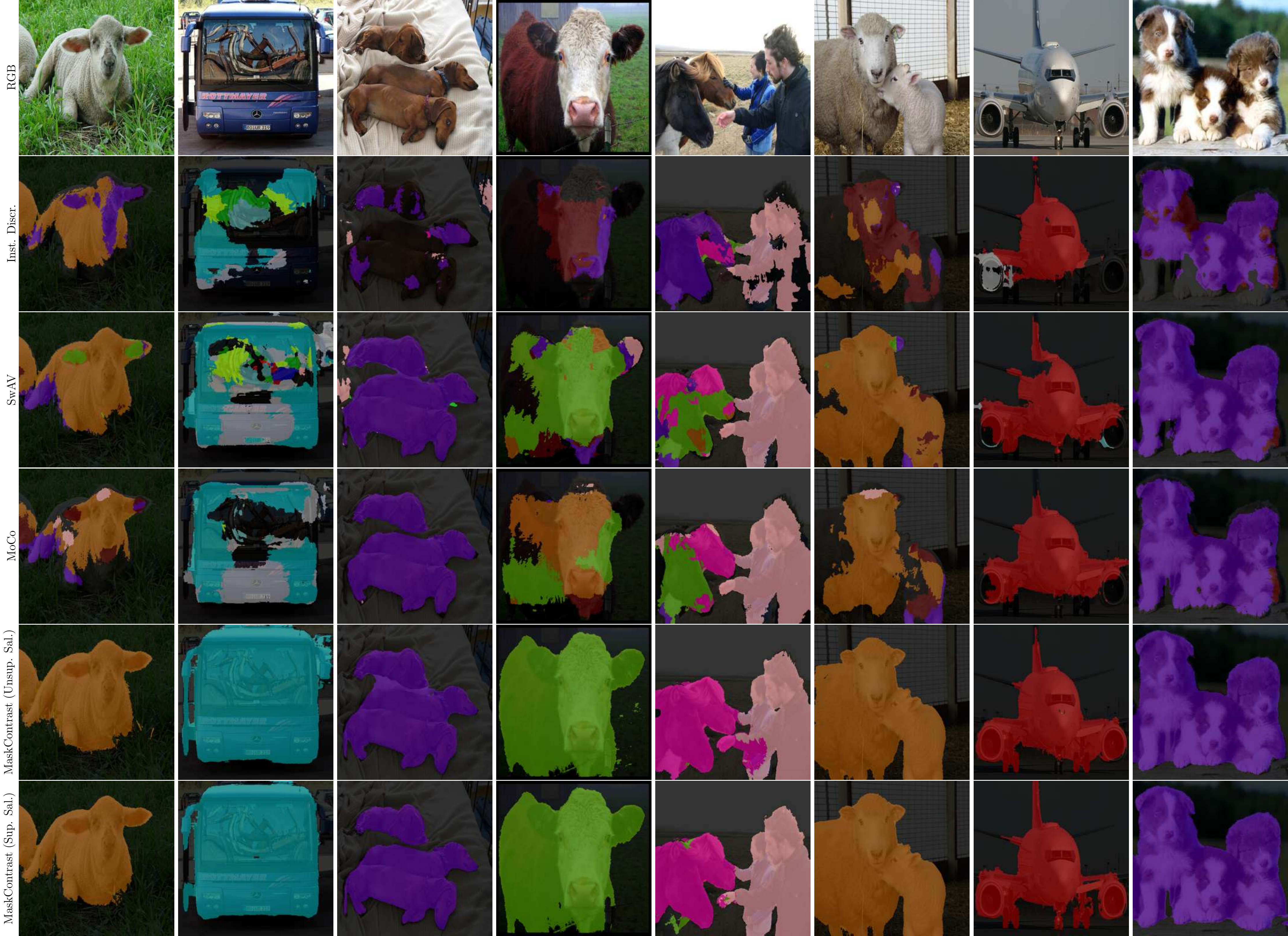}
    \caption{Qualitative comparison of the results after training a linear classifier on PASCAL. We use the MoCo weights to initialize our backbone.}
    \label{fig: linear_classifier}
\end{figure*}

\vspace{1em}

\noindent \textbf{Overclustering results.} K-Means does not employ any prior world knowledge, i.e. the ground-truth or target clusters are unknown. Therefore, it is unlikely that the predicted clusters will match the target ones on a complex and imbalanced dataset like PASCAL. To better understand the semantic structure discovered by the embedding space, we apply overclustering. In this case, a many-to-one mapping exists between the predicted and target clusters. Table~\ref{tab: overclustering} shows the results. The accuracy improves as we increase the number of predicted clusters. We hypothesize that local neighborhoods in the embedding space contain pixels of the same or visually similar objects, which benefits the performance when overclustering.
\begin{table}
    \tablestyle{4pt}{1.0}
    \begin{tabular}[t]{l|cc|cc}
    \textbf{Init.} & \multicolumn{2}{c}{MoCo v2} & \multicolumn{2}{c}{Sup.} \\
    \textbf{Sup. Sal.} & \xmark & \checkmark & \xmark & \checkmark \\
    \shline
    \textbf{Clusters} & & & & \\
    21 & 35.0 & 38.9 & 41.6 & 44.2 \\
    50 & 41.4 & 48.8 & 46.2 & 51.4 \\
    100 & 43.3 & 49.5 & 47.3 & 52.5 \\
    200 & 45.0 & 51.1 & 48.5 & 53.6 \\
    500 & 48.1 & 54.2 & 51.3 & 57.0 \\
    \end{tabular}
    \caption{Overclustering on PASCAL with MaskContrast (MIoU). We use MoCo or supervised ImageNet initial weights, and supervised (\checkmark) or unsupervised (\xmark) saliency.}
    \label{tab: overclustering}
\end{table}

\section{Semi-Supervised Learning}
\label{sec: semi_supervised}

This section describes the semi-supervised learning setup. In each case, we report the average result for three randomly sampled splits. 

\vspace{1em}
\noindent\textbf{ImageNet Pre-Trained Baseline.} We load the pre-trained ImageNet weights into a ResNet-50 backbone with dilated convolutions. We use batch size of 8 and stochastic gradient descent with momentum 0.9 and learning rate 0.004 in all data regimes. The learning rate was selected after performing a grid search. Additionally, we explored the use of different parameter groups with specific learning rate, e.g. the decoder used 10 times higher learning rate compared to the encoder. However, this did not result in any further improvements. We include a weight decay term $0.0001$. A poly learning rate scheduler is used. 

\vspace{1em}

\noindent\textbf{MaskContrast.} We use a batch size of 8 and learning rate of 0.004 when fine-tuning with 5\%, 12.5\% and 100\% of the labels. Differently, when using 1\% and 2\% of the labels, the learning rate is set to 0.001 for all layers in the network, except for the final convolutional layer which uses learning rate 0.1. The latter is well-motivated, since the complete network, including both encoder and decoder, were already pre-trained for the semantic segmentation task. The batch norm stats are frozen. We use stochastic gradient descent with momentum 0.9 and a weight decay term 0.0001. The learning rate is decayed using a poly learning rate scheduler. 

\section{Qualitative Results}

\label{sec: qualitative_results}
Figure~\ref{fig: linear_classifier} shows a qualitative comparison when training a linear classifier on top of the pre-trained representations. We compare the representations learned by our method using an unsupervised (5th row) or supervised (6th row) saliency estimator, against the ones from instance discrimination (2nd row)~\cite{wu2018unsupervised}, SWAV (3rd row)~\cite{caron2020unsupervised} and MoCo v2 (4th row) ~\cite{chen2020improved}. The qualitative results support the claim that our pixel embeddings learn semantically meaningful information. 